\documentclass[letterpaper, 10 pt, conference]{ieeeconf}  

\IEEEoverridecommandlockouts                              

\usepackage{graphics} 
\usepackage{epsfig} 
\usepackage{mathptmx} 
\usepackage{times} 
\usepackage{amsmath} 
\usepackage{amssymb}  

\title{\LARGE \bf
Fish-bone diagram of research issue: Gain a bird's-eye view on a specific research topic *
}

\author{JingHong Li$^{1}$, Huy Phan$^{2}$ Wen Gu$^{3}$, Koichi Ota$^{3}$ and Shinobu Hasegawa$^{3}$
\thanks{*This work was supported by JSPS KAKENHI Grant
Number JP20H04295.}
\thanks{$^{1}$Division of Advanced Science and Technology, Japan Advanced Institute of Science and Technology, Asahidai, Nomi, 9231292, Ishikawa, Japan.}%
\thanks{$^{2}$FPT University, HCMC, Vietnam, Department of Information Technology Specialization.}%
\thanks{$^{3}$Center for Innovative Distance Education and Research, Japan Advanced Institute of Science and Technology, Asahidai, Nomi, 9231292, Ishikawa, Japan.}%
}

\begin{document}

\maketitle
\thispagestyle{empty}
\pagestyle{empty}

\begin{abstract}

Novice researchers often face difficulties in understanding a multitude of academic papers and grasping the fundamentals of a new research field. To solve such problems, the knowledge graph supporting research survey is gradually being developed. Existing keyword-based knowledge graphs make it difficult for researchers to deeply understand abstract concepts. Meanwhile, novice researchers may find it difficult to use ChatGPT effectively for research surveys due to their limited understanding of the research field. Without the ability to ask proficient questions that align with key concepts, obtaining desired and accurate answers from this large language model \textit{(LLM)} could be inefficient. This study aims to help novice researchers by providing a fish-bone diagram that includes causal relationships, offering an overview of the research topic. The diagram is constructed using the issue ontology from academic papers, and it offers a broad, highly generalized perspective of the research field, based on relevance and logical factors. Furthermore, we evaluate the strengths and improvable points of the fish-bone diagram derived from this study's development pattern, emphasizing its potential as a viable tool for supporting research survey.

\end{abstract}


\section{Introduction}
\label{main}


In recent years, the digital environment for academic papers has improved significantly, with the rise of ChatGPT contributing to the development of text analysis tools. These tools use prompt engineering, text summarization, and question answering techniques to help researchers extract key information from large texts\cite{prompt-engineering}\cite{insight-survey}. However, understanding the basic concept can be challenging for researchers unfamiliar with a specific topic due to their lack of background knowledge\cite{VPRAS}. This difficulty can impede their ability to determine if their ideas have already been discussed and to identify the uniqueness of their research \cite{insight-survey}. In this case, it is essential to form a high-level overview of topics from a bird's-eye perspective. This requires more than just listing relevant tasks within the topic. It also involves linking related research issues across multiple papers and forming a succinct cause-and-effect relationship for novice researchers to grasp. Classifying an author's abstract in each article and creating a knowledge graph based on keywords can provide a certain overview of research domain \cite{PubMed}. However, since the abstract text does not include detailed information, such as 'issues in previous research' and 'issues the author addressed,' it is difficult to link it to prior tasks in that field and the significant issues in each task using only the abstract content.

On the other hand, most existing knowledge graphs offer a partial bird's eyes view, focusing on showcasing keyword relevance. This allows experts to quickly understand the key points\cite{KG}. However, for researchers unfamiliar with a particular research domain, the concept of a keyword map may be hard to understand. This is due to unclear extraction criteria and excessive conciseness, making it difficult to grasp an overview of the research topic. Thus, it becomes necessary to present high-dimensional knowledge structures in a logically structured way. For instance, computer science articles typically follow a specific pattern to declare research objectives : They start with historical context, analyze previous research issues, and then articulate the research purpose and contributions to improve the previous works. Therefore, creating an integrated graph to analyze this chain of research issues can highlight their correlation of issues across multiple articles. This approach could help novice researchers understand the key points of the research overview more logically. To address the issue mentioned above, we propose a new type of knowledge graph for a more comprehensive understanding of elements within a specific research topic: the fish-bone diagram , generated by an internal issue ontology, forming a high-dimensional, concise, and concrete summary. From the analysis, this fish-bone structure may help researchers conduct a more efficient bird's eyes view survey for their research.
The contributions of this study are as follows:

\textbf{1}. Define the issue ontology in academic papers, then manually annotate it at the sentence level to create a dataset.

\textbf{2}. A logical guide chain is established for novice researchers to navigate research topics. This chain, which expresses the research topic → task → issue ontology, serves as the foundation for in-depth data mining in a bird's eyes view survey.

\textbf{3}. Based on \textbf{2.}, we create a fish-bone diagram to provide a visual bird's eyes view of multiple academic papers.

\section{Related work}
In order to assist with research surveys, the previous work has focused on using important information from academic papers to generate automatic summaries or knowledge graphs. This involves (1) establishing criteria to identify important information and employing machine learning techniques to extract them from academic papers, and (2) developing approaches to link cause-and-effect relationships of essential information across various articles to improve readability.

Previous research that typically generates summarization by setting important information includes : Hayashi et al. introduce disentangled paper summarization, where models generate both contribution and context summaries simultaneously to hint at the contribution overview of a specific research direction\cite{contribution-summarizing}. Liu et al. expanded on this by introducing ContributionSum, a tool for automatically generating disentangled contributions\cite{ContributionSum}. This tool identifies author-written contributions in terms of Approach, Analysis, Result, Topic, and Resource. Moreover, integrating cited material from the contribution text can enhance the summarization by covering multiple papers ,such as Chen et al. propose the task of citation relation classification based on the contributions of cited papers to improve the summarization system for scientific papers\cite{Citation Relations}. These work commonly Summarize the part of contribution and related text that provide a road-map for the research survey. However, focusing solely on the contribution may overlook the source of research motivation — what gaps in previous studies prompted the author's contribution? Furthermore, covering the entire research topic comprehensively through the part of contribution alone is insufficient, which makes it difficult to understand the research overview. Conversely, previous research that generates a knowledge graph may better support understanding the research overview, such as Chan et al. proposed a representation ontology for an integrated four-space keyword based Knowledge Graph (background, objectives, solutions, and findings) using Natural Language Processing technology. However, the 'abstract' of the paper provides insufficient information to understand the detailed research issue fully. For instance, the part of 'abstract' seldom underscores the element of improvement from previous research, making it difficult to reflect the clue of author's motivation. Duan et al. generated summarization of the development of the research field over nearly a decade based on keywords, which played a key role for novice researchers to grasp the direction of the research field\cite{visualization technology}. However, the keyword-based knowledge graph might be too vague for novice researchers to understand the inherent logical connections among multiple academic papers. Zhang et al. present a heterogeneous network literature recommendation method based on the domain knowledge graph and hotspot information composition\cite{Hotspot Information Network}. However, relying solely on hotspot information is insufficient to delve into the historical issues remaining in the research domain and whether these issues have been resolved or not.

This study aims to expand the knowledge graph by introducing a new form of fish-bone diagram. Its main purpose is to give novice researchers a broad bird's-eye view of a research topic. This overview is constructed using issue ontology, reflecting the clue of the research task. By leveraging the logical structure of this issue ontology, we can effectively assist novice researchers in understanding the knowledge structure and motivation within a specific research topic.

\section{Methodology}
The methodology of this study is divided into two parts. First, we provide a detailed definition of Issue Ontology. Second, we introduce the fish-bone diagram, established based on Issue Ontology, to support research survey of bird's eyes view.

\begin{table*}[]
\centering
\caption{Fish-bone diagram configuration }
\label{fish-bone-t}
\begin{tabular}{lll}
\hline
\textbf{Part}        & \textbf{Expression}                  & \multicolumn{1}{c}{\textbf{Function}}                                                                                                                      \\ \hline
\textit{Head}        & Topic name                           & Present research topic.                                                                                                                                    \\ \hline
\textit{Joint}    & Tasks in Topic                       & \begin{tabular}[c]{@{}l@{}}Represent the 'effect' in a fish-bone diagram. \\ A research topic comprises multiple 'effects' (tasks).\end{tabular}            \\ \hline
\textit{Backbone}    & Issue Ontology in corresponding task & \begin{tabular}[c]{@{}l@{}}Express the label of 'emphasize issue’ and \\  'improvable issue’  presented in each task.\end{tabular}                                   \\ \hline
\textit{Fine-bones}   & Elements in Issue Ontology           & Summarize the in the clustered sentences of issue ontology                                                                                                \\ \hline
\textit{Child-bone} & Logic-link of issue summarization              & \begin{tabular}[c]{@{}l@{}}Express the logical chain composed of 'emphasize issue' \\ → 'improvable issue'.\end{tabular} \\ \hline
\end{tabular}
\end{table*}

\begin{figure*}[htbp]
\centering
\fbox{
\includegraphics[width=15.0cm,height=6.5cm]{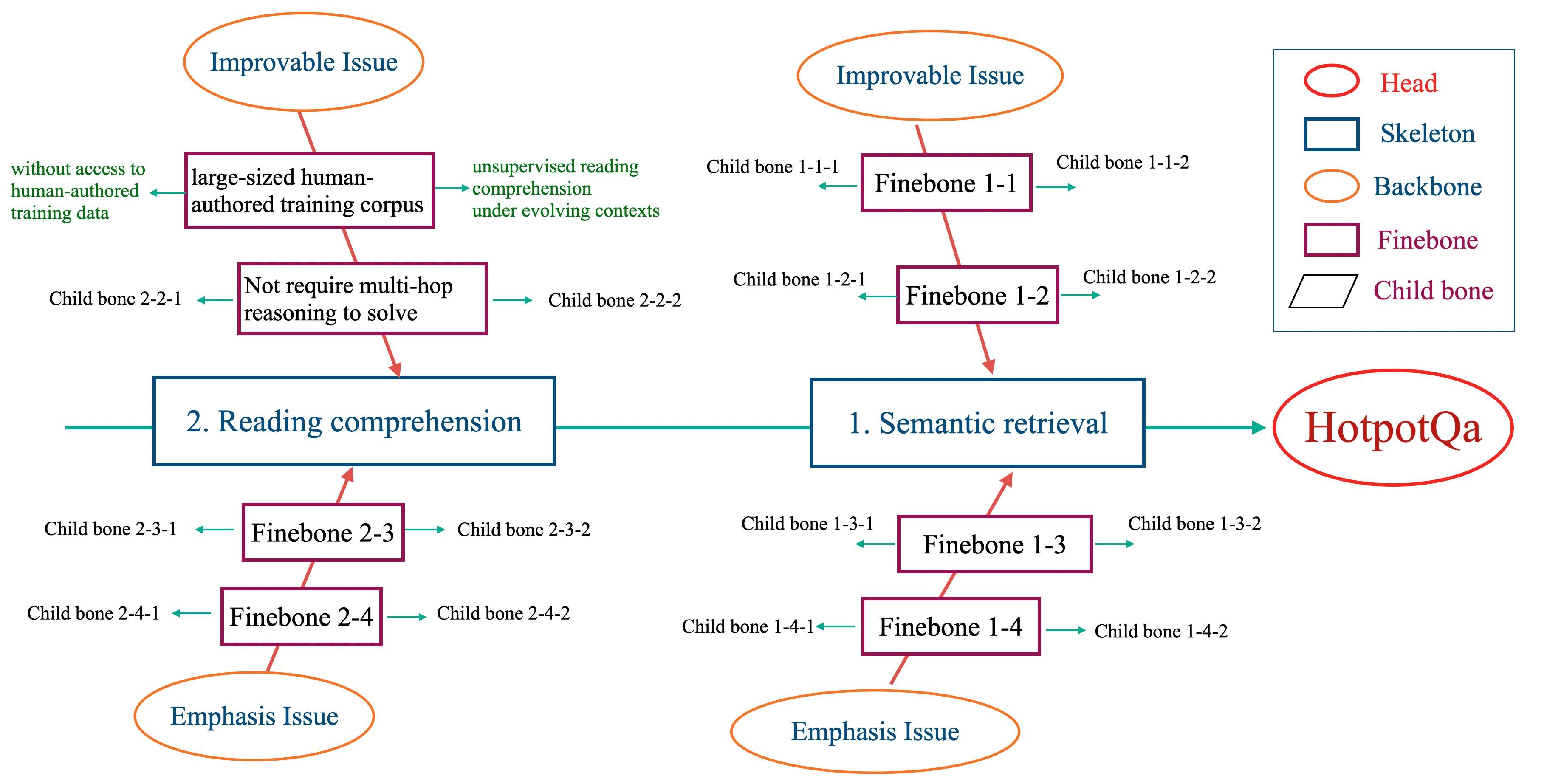}
}
\caption{Fish-bone diagram of bird’s eyes view survey}
\label{fish-bone}
\end{figure*}

\subsection{Bird's eyes view survey}
As birds can view a city's layout from the sky, including skyscrapers, open spaces, mountains, lakes, a similar view can be beneficial for novice researchers. Providing a high range covered bird's eyes view of the entire research topic helps researchers easily understand the logic of motivation to explore the direction from that research topic. 

\subsection{Mufti-modal Issue ontology in 'introduction'}
In academic papers, certain key elements provide insight into the author's reasoning and thought process. These elements reveal how the author discovers problems, investigates them, and contemplates solutions. This can be regarded as the writing clues of the article. The related sentences that appear throughout the papers make up the clues, which are crucial for readers to grasp the bird's eyes view of the research topic. These critical sentences encompass the concept of 'issue ontology'. Issues embody various debates within the academic world. Some of these have been resolved, some have been identified but remain unresolved, and others still require optimization and improvement. Their definitions are shown below:
\begin{itemize}
    \item 1. Prelude issue:  The root task mentioned in the paper reflects the historical context of this paper. It is usually in the head sentences of the introduction section as the source of clues.
    \item 2. Improvable issue: Some shortages the author mentioned from previous work that need to be solved.
    \item 3. Emphasize issue: Reflect on the author's purpose, contribution, and what they did.
\end{itemize}




\subsection{Design of fish-bone}
For novice researchers, merely summarizing an overview based on keywords or sentences may not clearly convey the logical structure, which could hinder knowledge management of the research topic. To address this, we adopt fish-bone diagram - causal diagrams created by Kaoru Ishikawa, are used to display the potential causes of a specific event\footnote{https://en.wikipedia.org/wiki/Ishikawa\_diagram}\footnote{https://en.wikipedia.org/wiki/Issue\_tree}. This type of diagram aids in understanding the learning logic from multiple articles. It includes preliminary tasks, challenges encountered, and the highlighted goals within a research topic. 
We present the issue ontology as a fish-bone diagram - a new knowledge graph for establishing the logical structure of issues in academic papers.

\subsubsection{Feature \& Element of fish-bone}
Fish-bone diagrams contain three key elements: effect, factor, and cause. By integrating these elements with issue ontology, the design of the fish-bone diagram can be expanded to include the feature of bird's-eye view, form the conceptual design of it as shown in Fig.\ref{fish-bone} and the structure summarized in TABLE \ref{fish-bone-t}.

Using the research topic of HotpotQA\cite{HotpotQA} as an example, it includes tasks like reading comprehension, semantic retrieval, and question answering. These tasks form the 'effect' of the fish-bone as joints. Each task encompasses an 'improvable issue,' which represents the persisting problems in the task. Meanwhile, 'emphasize issue' showcases the previously employed measures to tackle the 'improvable issue.' The ‘improbable issue’ and ‘emphasize issue’ constitute the backbone. Inside the backbone, the fine bone consists of sub-classes of its corresponding issue ontology, signifying the 'factor' in the fish-bone. For example, in the task of reading comprehension, there are some 'improvable issues' such as 'large-sized training corpus' and 'Do not require multi-hop reasoning to solve.' One method is called 'unsupervised reading comprehension' to solve this improvable issue of 'large-sized training corpus'. At this time, we call the 'emphasize issue' corresponding to this measure as child-bone. The child-bone and fine-bone with an objective logical relationship are connected by a cause-and-effect chain to form the 'cause' in the fish-bone. This provides researchers with an analysis path from task → Issue ontology → Issue logical chain, which helps researchers to capture bird's eyes view more quickly and establish connections between knowledge.

\section{Implementation}
\begin{figure*}[htbp]
\centering
\includegraphics[width=17.0cm,height=8.5cm]{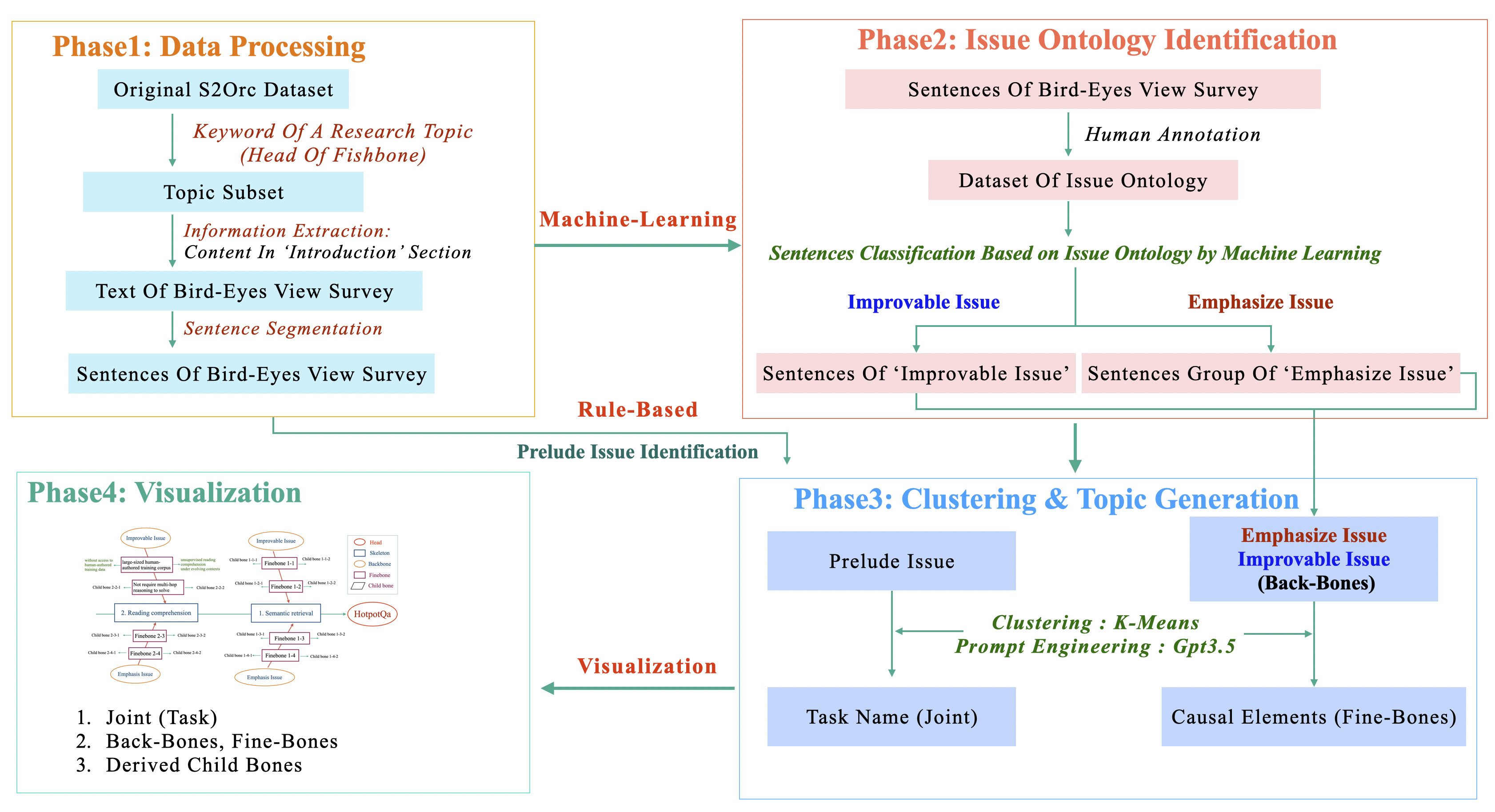}
\caption{Implementation procedure}
\label{figure_overview}
\end{figure*}

This chapter describes the process of developing a fish-bone diagram that provides a bird's-eye view of the HotpotQA research topic, based on the \textit{S2orc} dataset. The overview of implementation is shown in Fig.\ref{figure_overview}.
Table \ref{api and model} displays the APIs and models used at each stage of development.

\begin{table*}[]
\centering
\caption{API and model in implementation process }
\label{api and model}
\begin{tabular}{llll}
\hline
\textbf{Phase}                           & \textbf{API}     & \textbf{Method}                               & \textbf{Pre-training}            \\ \hline
\textit{Data processing}                 & \textit{SpaCy}   & \textit{Sentence Segmentation}                & \textit{en\_core\_sci\_lg}                \\ \hline
\textit{Issue ontology classification}   & \textit{Sklearn} & \textit{Sentence embedding, SVM, Grid search} & \textit{Distilbert-base-uncased} \\ \hline
Issue ontology clustering                & \textit{Sklearn} & \textit{Sentence embedding, K-means}          & \textit{Distilbert-base-uncased} \\ \hline
\textit{Task name generation}            & \textit{Chatgpt} & \textit{prompt engineering}                   & \textit{gpt-3.5-turbo}                   \\ \hline
\textit{Issue sentences summarization}   & \textit{Chatgpt} & \textit{prompt engineering}                   & \textit{gpt-3.5-turbo}                    \\ \hline
\textit{Fish-bone diagram visualization} & \textit{Pyvis}   & -                                             & -                                \\ \hline
\end{tabular}
\end{table*}

\subsection{Data processing}
The first step in constructing the bird's eyes view survey dataset is choosing a topic and filtering the relevant sub-dataset from the \textit{S2orc} dataset. We iterate through each paper in the \textit{S2orc} dataset, pulling out those that mention 'HotpotQA' for our sub-set. This sub-set comprises the full-text content and annotation-info of the papers linked with 'HotpotQA.' Next, we segment the text in 'introduction' section into sentences using \textit{'en\_core\_sci\_lg'}model from \textit{spaCy}\footnote{https://allenai.github.io/scispacy/}\cite{Spacy}  -  a comprehensive pipeline for biomedical data, including a 785k vocabulary and 600k word vectors. It provides impressive segmentation accuracy. Any complex segmentation patterns that failed were manually corrected.
\subsection{Issue ontology classification}

\subsubsection{Human annotation}
To create a machine learning dataset, experts annotated the sentence-implied issue ontology types based on the definitions provided in Section III.B. We strictly followed the rules detailed below during this annotation process. 

\textbf{(1)} In a few cases where two types of issue ontologies appear in one sentence, we manually split the sentence to ensure that each sentence carries only one type of issue ontology.

\textbf{(2)} If the author hypothesizes about a topic, it is also considered a contribution. Therefore, marked as an 'Emphasize issue.' Also, assessing the issue of a sentence in isolation is challenging, as the broader context of the targeted paper influences the annotation process.

\textbf{(3)} For sentences that do not fit our established issue ontology, we classify them as 'others' in our machine learning configuration. These sentences might include \textbf{1.} explanations of reasons, \textbf{2.} meaning of sentence do not match any of the three types of issue ontology, \textbf{3.} Issues have been addressed in previous research mentioned, and \textbf{4.} experimental results or performance achievements.

We found that the prelude issue typically emerges within the first 2-3 sentences of the 'introduction' section. Furthermore, the initial two sentences in most articles are usually sufficient to indicate the article's background task. Consequently, during the subsequent task clustering stage, we select the first two sentences from ‘introduction’ as the prelude issue for input.

\subsubsection{Result of issue ontology classification}
We utilize the \textit{'Distilbert-base-uncased'}\cite{dbu} pre-training model in \textit{Sentence BERT}\cite{sentence-bert} for sentence vectorization. This model was trained using resources like Wikipedia and BookCorpus. We chose support vector machine\textbf{\textit{(SVM)}} for classifying the vectorized sentences, as it demonstrates robust generalization performance. We also employ the grid search method\cite{GS} to identify the best parameters to apply. The datasets are shown in the TABLE \ref{detail dataset}
The classification results are shown in the TABLE \ref{result}, and the total classification accuracy has reached 78\%, which proves the effectiveness of the small-scale training data. However, since sentences with ‘emphasizing issue’ constitute a large portion of the collected articles, the imbalance in dataset labels may influence the classification results of ‘improvable issue.’

\begin{table}[htbp]
\centering
\caption{Detail of issue ontology dataset of bird's eyes view survey}
\label{detail dataset}
\begin{tabular}{|l|c|c|l|}
\hline
                          & \textbf{Train} & \textbf{Test} & \textbf{Total} \\ \hline
\textit{Emphasize issue}  & 239            & 77            & 316            \\ \hline
\textit{Improvable issue} & 139            & 48            & 187            \\ \hline
\textit{Others}           & 214            & 73           &  287            \\ \hline
\textit{Total}            & 592            & 198           & 790           \\ \hline
\end{tabular}
\end{table}

\begin{table}
\centering
\caption{Classification result of issue ontology}
\label{result}
\begin{tabular}{|l|c|c|c|}
\hline
                  & \multicolumn{1}{l|}{\textbf{Precision}} & \multicolumn{1}{l|}{\textbf{Recall}} & \multicolumn{1}{l|}{\textbf{F1-score}} \\ \hline
\textit{Emphasize issue}   & 0.92                                   & 0.76                               & 0.84                                  \\ \hline
\textit{Improvable issue}  & 0.65                                    & 0.79                                 & 0.71                                   \\ \hline
\textit{Others} & 0.73                                    & 0.80                                 & 0.76                                   \\ \hline
\end{tabular}
\end{table}

\subsection{Fish-bone diagram}
We configure the fish-bone diagram to approach Figure \ref{fish-bone} following the description in Table \ref{fish-bone-t}.

\subsubsection{Joint: Prelude issue clustering \& task name generation}

In this section, we introduce the process of using prelude issue ontology content to construct the joint (effect) of fish-bone. We start by selecting the first two sentences from the 'introduction' section of each article to serve as the prelude issue. For sentence vector embedding, we also use the 'distilbert-base-uncased' model from \textit{Sentence BERT}. These sentence vectors are then clustered using the k-means\cite{K-means} method. Finally, we generate the task name for each category via simple prompt engineering by the description ‘ Find a theme for the following text, and the generated theme is limited to within 5 words.’ These tasks form the joint (effect) component of the fish-bone.

\subsubsection{Back-bone\& Fine-bone}
We extend \verb|n| branches from the task's joint to identify factors impacting each task. These branches represent the 'Emphasize' and 'Improvable' issues associated with the task. Each branch is created using the results of the set of \textit{Sentence BERT} + issue cluster by k-means, and prompt engineering description ‘Your task is to find n themes for the following text, Limit each theme to 5 words’ , collectively referred to as 'fine-bones'. Here n represents the number of fine-bone to be generated. These 'fine-bones', composed of issue ontologies, significantly influence tasks, which is why we refer to them as contributing factors.

\subsubsection{Child-bone}
We build the child-bone, the most basic unit of the fish-bone design, derived from the fine-bone. This is done using the 'Improvable issue' ←→ 'Emphasize issue' logic chain within the research task. These logic connections diverge and radiate based on a cluster of real connections of the issue ontology. 
Specifically, we link the ‘emphasize issue‘ that appears in the article corresponding to each fine-bone to the ’improvable issue‘ that appears in the same article, and then use the prompt engineering to generate a brief summary of the sentence groups of 'improvable issue', completing the connection between the issue summaries. The process is the same when shifting from  'improvable issue' to 'emphasized issue.'


\section{Conclusion}
This study focused on automatically generating bird's-eye view fish-bone diagrams for research topics to assist novice researchers. This process uses issue ontology units, logically organized and expanded to generate the diagrams. We started by collecting introductory text from academic papers related to a specific research topic, which was then segmented into sentences level. Expert researchers annotated these sentences according to the implied issue ontology type, forming the training dataset for the bird's-eye view survey. Next, we utilized rule-based and machine learning methods to categorize and extract sentences related to prelude, improvable, and emphasized issues. We summarize the prelude issues to form the tasks of the diagram using clustering and prompt engineering. The backbone and fine-bones, which illustrate the cause-and-effect relationship, were created from the summarized ‘emphasized issues’ and ‘improvable issues’ by issue ontology classification, clustering and prompt engineering. Subsequently, we generated the child-bone from the objectives logical of ‘emphasize issue’ ←→‘improvable issues’. Lastly, we evaluated the readability of the generated diagram and the sustainability of the development through case analyses. For the future expansion and improvement of this study, the following points are proposed:\\
\textbf{(1) }Expanding the issue ontology: This study focused on automatically generating a fish-bone diagram from the 'introduction' sections of academic papers. However, issues discussed in other sections, like addressed and improvable issues in the 'related work' section, or resolved and finding issues in the 'conclusion' section, are not covered. These issue ontologies can also highlight key research points, such as problems addressed across multiple articles and future research directions. Hence, identifying these types of issue ontologies is a task for future work.\\
\textbf{(2) }Analysis relevance of novelty: By analyzing the Relevance of Novelty: We can study the novelty's relevance by combining addressed, improvable, and emphasized issues. For example, we can consider similar addressed or improvable issues across multiple articles with different emphasized issues to implement originality analysis, which can provide additional insights for researchers.\\

\end{document}